\documentclass[10pt,twocolumn,letterpaper]{article}

\usepackage[pagenumbers]{cvpr} 

\usepackage{amsmath}  
\usepackage{amsthm}  
\usepackage{graphicx}  
\usepackage{tcolorbox} 
\usepackage{multirow}
\usepackage{colortbl}
\usepackage{bbding}
\usepackage{wrapfig}
\usepackage{array}  

\usepackage{amssymb,booktabs}
\usepackage{tabularx}
\usepackage{bbding}
\usepackage{pifont}
\usepackage{makecell}
\usepackage{ulem}

\definecolor{mymutedblue}{RGB}{222,236,250}

\definecolor{cvprblue}{rgb}{0.21,0.49,0.74}
\usepackage[pagebackref,breaklinks,colorlinks,allcolors=cvprblue]{hyperref}

\def\confName{CVPR}
\def\confYear{2026}

\title{RemoteShield: Enable Robust Multimodal Large Language Models for Earth Observation}

\author{
\textbf{Rui Min}\textsuperscript{1,*},
\textbf{Liang Yao}\textsuperscript{1,*},
\textbf{Shiyu Miao}\textsuperscript{2,*},
\textbf{Shengxiang Xu}\textsuperscript{3}\\
\textbf{Yuxuan Liu}\textsuperscript{1},
\textbf{Chuanyi Zhang}\textsuperscript{1},
\textbf{Shimin Di}\textsuperscript{3},
\textbf{Fan Liu}\textsuperscript{1,†}
\\
\textsuperscript{1}Hohai University \quad
\textsuperscript{2}Nanjing University \quad
\textsuperscript{3}Southeast University \\
\\
{\tt\small  \textsuperscript{*}Equal Contribution \quad
\textsuperscript{†}Corresponding Author}
\\
{\tt\small Email: fanliu@hhu.edu.cn}
\\
{\tt\small GitHub Repo: \url{https://github.com/SteveJoker404/RemoteShield}}
}

\begin{document}
\maketitle
\begin{abstract}
A robust Multimodal Large Language Model (MLLM) for Earth Observation should maintain consistent interpretation and reasoning under realistic input variations. However, current Remote Sensing MLLMs fail to meet this requirement. Trained on carefully curated clean datasets, they learn brittle mappings that do not generalize to noisy conditions in operational Earth Observation. Consequently, their performance degrades when confronted with imperfect inputs in deployment. To quantify this vulnerability, we construct a realistic set of multimodal perturbations, including visual degradations such as cloud and fog cover, together with diverse human-centric textual variations ranging from colloquialisms to vague or omitted instructions. Empirical evaluations show that these perturbations significantly impair the visual-semantic reasoning capabilities of leading RS foundation models. To address this limitation, we introduce RemoteShield, a robust Remote Sensing MLLM trained to maintain consistent outputs across realistic input variations. During training, each clean sample is paired with its image-text perturbed variants to form a semantic equivalence cluster. Rather than directly fitting noisy samples, RemoteShield is optimized through preference learning over clean and perturbed conditions within the same cluster. By comparing model responses to clean and corrupted inputs, the model is encouraged to favor stable responses over perturbation-induced failures. This cross-condition alignment helps the model focus on underlying task semantics despite visual degradations and textual noise. Experiments on three Earth Observation tasks show that RemoteShield consistently delivers stronger robustness and cross-condition consistency than representative baselines under realistic multimodal perturbations.

\end{abstract}    
\section{Introduction}

\begin{figure}[t]
    \centering
    \includegraphics[width=0.97\columnwidth]{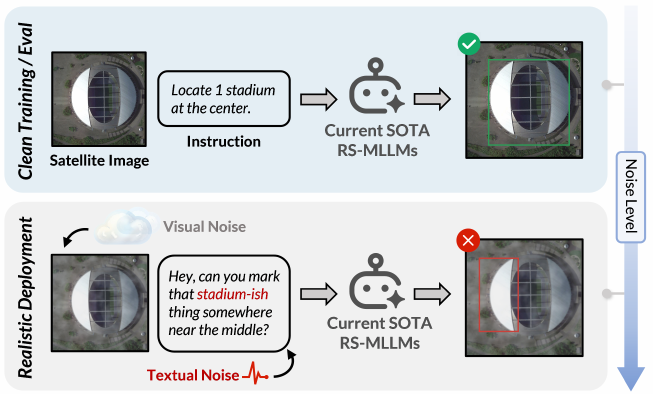}
    \caption{\textit{Performance Collapse under Real-World Noise.} (1) With clean image-text inputs, current RS-MLLMs can complete the localization task. (2) When realistic visual degradation and colloquial query noise are introduced, the same task may fail even though the underlying intent remains unchanged.}
    \label{fig:fig1}
\end{figure}

A robust Multimodal Large Language Model (MLLM) for Earth Observation (EO)~\cite{kansakar2016review,binci2026earth,connors2025earth} is essential for reliable real-world deployment.
As MLLMs are increasingly used for remote sensing vision-language tasks~\cite{zhou2025advances,zhu2026foundations,huang2025survey,zhou2024towards}, maintaining consistent performance under unconstrained environmental conditions becomes a fundamental requirement~\cite{chou2025mm,chou2026test,rabanser2026towards}. In real-world EO applications, inputs are often degraded or reformulated rather than matching the clean conditions of standard benchmarks~\cite{robust_bench_1, robust_bench_2, robust_bench_3, reobench}.
On the visual side, optical satellite imagery is often degraded by atmospheric interference such as cloud cover and fog~\cite{shen2014effective,long2013single}. On the textual side, human queries often introduce significant noise due to diverse user backgrounds, manifesting as colloquial expressions, verbose descriptions, or vague instructions~\cite{liang2026prompt,dumpala2024sugarcrepe++,shah2025analyzing}. Consequently, a practical and robust EO MLLM is required to yield accurate and consistent predictions, regardless of whether it processes meticulously curated high-quality image-text pairs or their severely degraded counterparts typical of actual deployment.

As shown in Figure~\ref{fig:fig1}, while current RS MLLMs~\cite{luo2025geoevolve,yao2025remotereasoner,wu2026vision,yao2026remoteagent} exhibit strong visual-semantic understanding on standard benchmarks and effectively align fine-grained spatial features with complex textual semantics under controlled conditions, they remain vulnerable under realistic deployment conditions. This vulnerability is closely tied to current instruction-tuning practices~\cite{wei2021finetuned,sanh2021multitask,ouyang2022training,wang2023self} and their reliance on curated supervision.
To achieve precise cross-modal alignment, existing foundation models are largely trained on curated datasets~\cite{zhang2024earthgpt,li2024vrsbench,directsam_rs,yao2025falcon,remotesam} with limited noise.
Consequently, the models tend to establish cross-modal correspondences between visual tokens and text embeddings that are finely tuned to high-quality inputs. Since complex multimodal perturbations are typically underrepresented in the training distribution, the resulting representation spaces can remain sensitive to out-of-distribution interference. When confronted with deployment-time noise, the visual encoder may struggle to process degraded spatial features, while the language backbone often experiences reduced capability in parsing highly colloquial or vague instructions.

To empirically quantify the aforementioned vulnerabilities, we construct a comprehensive evaluation dataset containing realistic multimodal perturbations. Specifically, we systematically corrupt standard image-text pairs by injecting the previously discussed atmospheric interference into the visual inputs, and reformulating formal task instructions into diverse, human-centric formats. We then evaluate multiple representative remote sensing multimodal large language models using these perturbed samples. The empirical results demonstrate that the introduction of these coupled multimodal perturbations leads to a substantial and consistent decline in performance across the evaluated models. This quantitative evidence confirms the limitation of existing foundation models when processing unconstrained, noisy inputs. These results suggest that the core issue is not merely insufficient exposure to noisy inputs, but the lack of an explicit training signal that encourages stable behavior across semantically equivalent input conditions.

Motivated by this observation, we propose RemoteShield, whose primary objective is to maintain consistent outputs across varying input conditions, rather than being trained merely by fitting noisy samples. Specifically, we organize a clean sample and its corresponding image-text perturbed variants into a unified semantic equivalence cluster. Instead of applying supervised fine-tuning (SFT)~\cite{zhang2026instruction} to these noisy inputs, RemoteShield is trained with preference optimization~\cite{rafailov2023direct}. By comparing the model's generated responses to clean and degraded inputs within the same cluster, this method explicitly encourages the model to separate robust responses from perturbation-induced failures. This cross-condition alignment trains the model to extract the shared semantic information across diverse perturbations, thereby preserving stable behavior despite visual degradations and textual noise.

Experiments on scene classification, visual question answering, and visual grounding show that RemoteShield consistently outperforms representative baselines under realistic multimodal perturbations, yielding stronger robustness and more stable outputs across semantically equivalent conditions. These gains are consistent across both general-domain and remote-sensing-specific baselines. Ablations and further analyses further verify the effectiveness of the proposed perturbation-driven preference construction and preference-based alignment strategy.

The main contributions of this work are summarized as follows:

\begin{itemize}
    \item We construct a comprehensive multimodal perturbation dataset that simulates realistic environmental visual degradations and diverse forms of human-centric textual noise in real-world Earth Observation scenarios.
    \item We evaluate a set of representative remote sensing MLLMs, quantitatively revealing their inherent limitations and significant performance degradation when processing unconstrained noisy inputs in practice.
    \item We propose RemoteShield, whose training framework shifts robustness learning from passive noise fitting to active cross-condition alignment. It preserves stable reasoning capabilities under varying input quality conditions.
    \item Extensive experiments demonstrate that RemoteShield effectively mitigates the impact of coupled multimodal perturbations, improving the accuracy and reasoning consistency of the model in noisy environments.
\end{itemize}
\section{Related Work}

\subsection{Earth Observation MLLMs}

Earth Observation MLLMs have progressively developed from EO-specific data construction to broader geospatial assistance~\cite{wang2024earthvqa,jiang2025eaglevision,xue2024reo,liu2024rsunivlm,zhang2024earthmarker}. Early efforts such as RSGPT and SkySenseGPT established this direction through EO-aligned image-text resources, large-scale instruction tuning, and dedicated benchmarks for remote sensing vision-language learning~\cite{hu2025rsgpt,luo2024skysensegpt}. Building on this foundation, later models further expanded beyond image-level understanding: RingMoGPT moved toward unified modeling across scene understanding, grounding, and change analysis~\cite{wang2024ringmogpt}, while TEOChat extended EO assistants to complex temporal observation sequences~\cite{irvin2024teochat}.

More recently, the focus has begun to shift toward more explicit reasoning and finer spatial understanding in EO settings~\cite{zhang2025geo,wang2025geozero,li2026georeason,fiaz2025geovlm,shabbir2025thinkgeo}. Geo-CoT emphasizes grounded geospatial reasoning~\cite{liu2025towards}, and SegEarth-R1 further pushes EO MLLMs toward pixel-level reasoning under implicit user queries~\cite{li2025segearth}. In parallel, task-specific EO vision-language studies have extended this landscape from change-aware question answering and grounding~\cite{li2024show} to more precise and open-vocabulary visual grounding in remote sensing imagery~\cite{li2024language,li2026provg,li2026rsvg}. Despite this progress, existing EO MLLMs are still developed mainly to expand capability and improve benchmark performance under carefully curated conditions. Their robustness and behavioral stability under realistic multimodal perturbations remain underexplored.

\subsection{Robustness of Multimodal Models}

Robustness is increasingly important as evaluation moves beyond curated benchmarks toward more realistic deployment settings~\cite{saxena2026vlm}. Prior studies show that vision-language models and multimodal large language models can be vulnerable to distribution shifts and multimodal perturbations, indicating that strong clean performance alone does not guarantee reliable behavior in practice~\cite{chen2023benchmarking,zhao2023evaluating}. This has motivated a growing body of work on robustness assessment, including benchmark construction, corruption-based evaluation, and broader studies of multimodal reliability~\cite{jiang2025survey}.

Recent efforts further examine this fragility from both visual and linguistic perspectives~\cite{saxena2026vlm}. On the visual side, image corruptions and degradations substantially affect multimodal reasoning, with different corruption types inducing distinct robustness patterns across tasks and model families~\cite{qiu2025benchmarking,usama2025analysing}. On the language side, prompt sensitivity and textual variation also introduce substantial instability, showing that multimodal predictions may change even under semantically equivalent input formulations~\cite{liang2026prompt,alhamoud2025vision}. A smaller body of work has begun to study robustness through consistency and stable behavior across perturbed conditions, rather than accuracy alone, and to explore optimization strategies for improving invariance under semantic variation~\cite{rabanser2026towards,xu2025robustflow}. However, these efforts are still centered on general-domain multimodal models. In contrast, EO MLLMs remain rarely studied under realistic coupled image-text perturbations, especially from the perspective of cross-condition consistency and behavioral stability.

\begin{figure*}[t]
    \centering
    \includegraphics[width=1\textwidth]{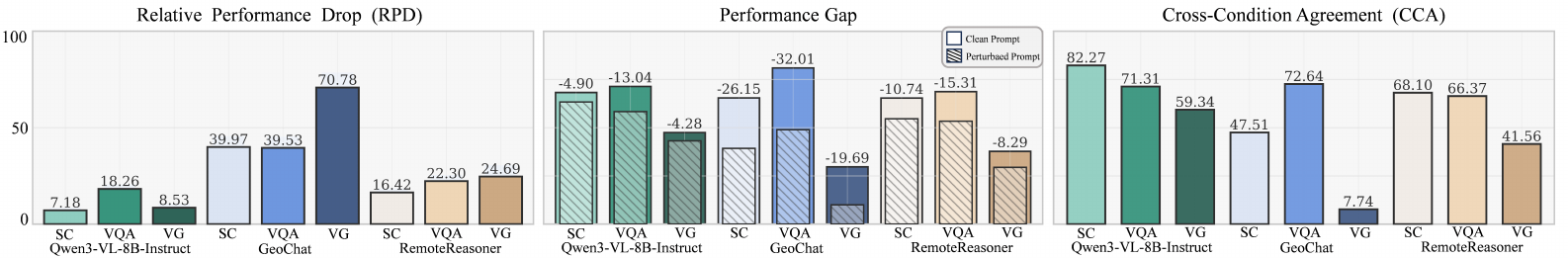}
    \caption{Motivational diagnosis of representative baseline MLLMs under the proposed perturbation setting. The left, middle, and right panels show Relative Performance Drop (RPD), Performance Gap $\Delta M = M_{\mathrm{pert}} - M_{\mathrm{clean}}$, and Cross-Condition Agreement (CCA), respectively. SC, VQA, and VG denote \textit{scene classification}, \textit{visual question answering}, and \textit{visual grounding}.}
    \label{fig:motivation}
\end{figure*}
\section{Robustness Evaluation}
\label{sec:robustness_evaluation}

Clean benchmark performance alone does not determine whether an RS-MLLM remains reliable when the same EO task is presented with degraded imagery and irregular yet semantically equivalent queries. We therefore construct matched clean--perturbed image-query pairs and use them to build the datasets in this work. On this basis, we introduce robustness and cross-condition consistency metrics to measure performance degradation under perturbation and output stability across semantically equivalent inputs. Evaluating representative baselines under this protocol reveals substantial robustness gaps, motivating an RS-MLLM that remains both accurate and behaviorally stable under multimodal perturbations.

\subsection{Multimodal Perturbation}
\label{sec:multimodal_perturbation}

To evaluate robustness under semantically equivalent input variation, we perturb EO image-query pairs while keeping the underlying task semantics unchanged. Textual perturbations mimic realistic human query variation, whereas visual perturbations simulate EO-specific image degradation. Together, they create matched clean--perturbed pairs that expose whether a model can preserve stable behavior when both modalities vary in realistic ways.

\subsubsection{Text Perturbation}

On the textual side, we aim to simulate realistic human query variation while preserving task semantics. In practical EO applications, queries are often colloquial, redundant, fragmented, or context-dependent, even though the underlying intent remains unchanged. To model this variation, we use \textbf{Qwen3.5-27B} to rewrite each query under prompts that preserve semantic anchors while changing the surface form. We consider four perturbation regimes that capture representative forms of human-centric textual variation:

\begin{itemize}
    \item \textbf{Naturalistic.} Casual, mildly irregular task expressions with spoken-style phrasing and occasional self-correction.
    \item \textbf{Conversational.} Longer and less direct requests with clarification, backtracking, or task-related elaboration.
    \item \textbf{Shorthand-notes.} Compressed query fragments with omitted function words and keyword-heavy phrasing.
    \item \textbf{Persona.} Context-dependent reformulations shaped by urgency, operational pressure, or role-specific language.
\end{itemize}
 The complete prompting templates and generation settings are provided in the supplementary material. Because LLM rewriting is stochastic, some outputs may still exhibit semantic drift or unnatural phrasing; we therefore manually verify all rewritten queries and revise problematic cases when necessary.

\subsubsection{Image Perturbation}

On the visual side, we simulate atmospheric degradation common in real-world EO imagery. Rather than applying generic random corruption, we model structured cloud--fog interference together with visibility loss, which weakens fine-grained spatial cues while preserving the underlying scene content and task target.

Given an input image $I \in [0,1]^{H \times W \times 3}$ and a perturbation strength $s \in [0,1]$, we define the visual perturbation operator as
\begin{equation}
I' = \mathcal{T}_I(I,s),
\label{eq:image_perturb_final}
\end{equation}
where $\mathcal{T}_I$ synthesizes a low-frequency cloud mask from multi-octave random noise, blends it with a fog-colored veil, and then applies contrast attenuation, brightness lifting, and resolution-aware Gaussian blur. In this way, the single scalar $s$ jointly controls cloud coverage, fog opacity, and clarity loss. Full implementation details are deferred to the supplementary material.

\subsection{Benchmark Construction}
\label{sec:dataset_construction}

Using the perturbation rules above, we build matched clean and perturbed datasets for training and evaluation. For each source set, samples are partitioned into four disjoint subsets, each assigned to one textual perturbation regime, while images are perturbed with a fixed visual strength of $0.45$.

For training, we use \textit{GeoChat-Instruct}~\cite{kuckreja2024geochat} as the clean source set and retain about 20,000 samples associated with more than 14,000 remote sensing images. The corpus covers scene classification, visual grounding, and visual question answering (VQA), and the original supervision target is preserved under perturbation.

For evaluation, we construct task-specific clean and perturbed test sets from three source datasets: 3,000 scene classification examples from AID~\cite{xia2017aid}, approximately 7,000 VQA examples from the test split of \textit{RSVQA-LRBEN}~\cite{lobry2020rsvqa}, and 2,000 visual grounding examples from the \texttt{refer} and \texttt{grounding} subsets of \textit{GeoChat-Bench}~\cite{kuckreja2024geochat}. In each case, the clean test set is perturbed with the same protocol to form matched clean--perturbed pairs.

\begin{figure*}[!t]
    \centering
    \includegraphics[width=\textwidth]{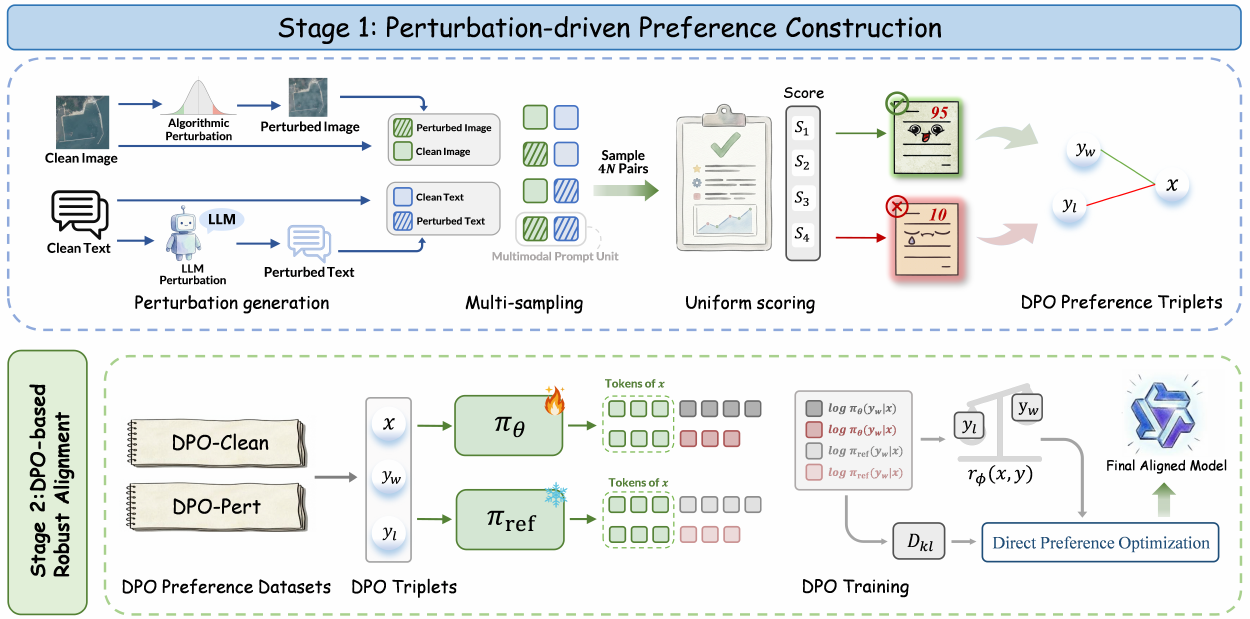}
    \caption{Overview of the RemoteShield training framework. Stage 1 constructs the DPO training triplets through multi-condition response inference, unified quality scoring, and preference pair selection over semantically equivalent multimodal inputs. Stage 2 performs robust alignment via Direct Preference Optimization (DPO) using the resulting preference triplets.}
    \label{fig:overview}
\end{figure*}

\subsection{Robustness Metrics}
\label{sec:robustness_consistency_evaluation}

Given matched clean and perturbed inputs, our evaluation focuses on two questions: how much task performance deteriorates under perturbation, and whether model behavior remains stable across semantically equivalent conditions. Standard task metrics reflect task success, but they do not capture relative degradation or cross-condition stability. We therefore report the standard task metrics together with Relative Performance Drop (RPD) for robustness and Cross-Condition Agreement (CCA) for consistency.

For scene classification and visual question answering, we use accuracy as the metric. For visual grounding, we report Acc@0.5 and gIoU. We compute them under deterministic decoding.

RPD quantifies the relative loss in task performance under perturbation:

\begin{equation}
\mathrm{RPD}
=
\frac{M_{\mathrm{clean}}-M_{\mathrm{pert}}}{M_{\mathrm{clean}}}\times 100\%,
\end{equation}
where $M_{\mathrm{clean}}$ and $M_{\mathrm{pert}}$ denote performance under the clean and perturbed conditions, respectively. In our experiments, $M$ corresponds to accuracy for scene classification and VQA, and to Acc@0.5 for visual grounding. Smaller RPD indicates stronger robustness.

CCA is computed from stochastic decoding. For each evaluation example, we draw $K=5$ outputs under both the clean and perturbed conditions using different random seeds. For text-valued tasks such as scene classification and VQA, let $Y_i^c$ and $Y_i^p$ denote the sampled outputs of the $i$-th example under the clean and perturbed conditions, respectively, and let $\operatorname{mode}(\cdot)$ denote the most frequent prediction in a sampled output group. We then define
\begin{equation}
\mathrm{CCA}^{\mathrm{text}}
=
\frac{1}{N}\sum_{i=1}^{N}
\mathbb{I}\!\left[
\operatorname{mode}(Y_i^c)=\operatorname{mode}(Y_i^p)
\right],
\end{equation}
where $\mathbb{I}[\cdot]$ is the indicator function.

For visual grounding, exact mode matching is not applicable because the outputs are continuous bounding boxes. Let $\mathcal{B}_i^c=\{B_{i,1}^c,\dots,B_{i,K}^c\}$ and $\mathcal{B}_i^p=\{B_{i,1}^p,\dots,B_{i,K}^p\}$ denote the sampled box sets under the clean and perturbed conditions. We therefore define
\begin{equation}
\resizebox{0.9\linewidth}{!}{%
$\displaystyle
\mathrm{CCA}^{\mathrm{vg}}
=
\frac{1}{N}\sum_{i=1}^{N}
\frac{1}{K^2}
\sum_{m=1}^{K}\sum_{n=1}^{K}
\mathrm{IoU}_{\mathrm{match}}(B_{i,m}^c,B_{i,n}^p),
$%
}
\end{equation}
where $\mathrm{IoU}_{\mathrm{match}}(\cdot,\cdot)$ denotes the Hungarian-matched IoU between two sampled box sets. Higher CCA indicates stronger stability.

Taken together, RPD and CCA complement the standard task metrics by quantifying robustness and behavioral stability under semantically equivalent multimodal perturbations.

\subsection{Evaluation Results}
\label{sec:motivational_analysis}

With the perturbation benchmark and robustness metrics in place, we examine whether current RS-MLLMs remain reliable under semantically equivalent but degraded inputs. As shown in Fig.~\ref{fig:motivation}, degradation is systematic across tasks. Even the strongest general-domain baseline, Qwen3-VL-8B-Instruct, exhibits clear robustness loss, with RPDs of 7.18 on scene classification, 18.26 on VQA, and 8.53 on visual grounding. RS-specialized models are often more brittle: GeoChat degrades severely on visual grounding, reaching an RPD of 70.78 and a CCA of only 7.74, while RemoteReasoner remains more stable but still shows only moderate cross-condition agreement. The performance-gap panel confirms these relative drops correspond to substantial absolute loss under perturbation.

These results reveal a limitation that goes beyond ordinary performance degradation under noise. Current RS-MLLMs do not merely become less accurate when the input is corrupted, they also fail to preserve stable behavior when different input forms express the same underlying task semantics. Strong clean performance therefore does not imply perturbation-invariant performance. What is missing is an explicit training signal that enforces consistent decisions across matched clean and perturbed conditions, allowing the model to preserve both task success and behavioral stability under realistic multimodal perturbations.

\section{RemoteShield}
\label{sec:method}

The robustness evaluation above shows that current RS-MLLMs often suffer not only from degraded task performance under perturbation, but also from unstable behavior across semantically equivalent input conditions. To address this limitation, we propose RemoteShield, a robust RS-MLLM designed to maintain consistent task behavior across semantically equivalent clean and perturbed inputs. As illustrated in Fig.~\ref{fig:overview}, RemoteShield follows a two-stage training framework: it first constructs perturbation-driven preference triplets from matched clean and perturbed inputs, and then aligns the policy via Direct Preference Optimization (DPO)~\cite{rafailov2023direct}.

\subsection{Formulation}

Building directly on the perturbation setting above, RemoteShield treats a clean EO instance and its perturbation-derived variants as a unified set of semantically equivalent input conditions. Concretely, for a clean image-query pair $(I_i,q_i)$ and the corresponding perturbed image and query $(I_i',q_i')$, we define

\begin{equation}
\resizebox{0.9\linewidth}{!}{%
$\displaystyle
\mathcal{X}_i =
\left\{
x_i^{(1)}=(I_i,q_i),\;
x_i^{(2)}=(I_i',q_i),\;
x_i^{(3)}=(I_i,q_i'),\;
x_i^{(4)}=(I_i',q_i')
\right\},
$%
}
\label{eq:condition_set}
\end{equation}
which correspond to the clean, image-only perturbed, text-only perturbed, and jointly perturbed conditions, respectively. Although these conditions differ in form, they share the same semantics and therefore the same reference target, denoted by $y_i^\star$.

Each condition $x_i^{(j)} \in \mathcal{X}_i$ corresponds to the same underlying task instance and should therefore induce semantically consistent model behavior. During data construction, we use the initialization policy $\pi_{\mathrm{base}}$ to sample multiple candidate responses from these matched input conditions, and later optimize the aligned policy $\pi_\theta$ using the resulting preference data.

\begin{table}[t]
\centering
\caption{Main results on scene classification.}
\label{tab:scene_classification}
\resizebox{0.98\columnwidth}{!}{%
\begin{tabular}{lcccc}
\toprule
Method & Clean Acc. & Pert Acc. & RPD $\downarrow$ & CCA $\uparrow$ \\
\midrule
\rowcolor{gray!30}
\multicolumn{5}{l}{\textbf{\textit{General-domain}}} \\
LLaVA-1.5-7B~\cite{liu2024improved} & 48.13 & 43.40 & 9.83 & 62.93 \\
Qwen3-VL-8B-Instruct~\cite{bai2025qwen3} & \underline{68.23} & \underline{63.33} & \underline{7.18} & \underline{82.27} \\
\rowcolor{gray!30}
\multicolumn{5}{l}{\textbf{\textit{RS-Specialized}}} \\
GeoChat~\cite{kuckreja2024geochat} & 65.43 & 39.28 & 39.97 & 47.51 \\
RemoteReasoner~\cite{yao2025remotereasoner} & 65.37 & 54.63 & 16.42 & 68.10 \\
\rowcolor{mymutedblue}
\multicolumn{1}{>{\columncolor{mymutedblue}}l}{RemoteShield} &
\multicolumn{1}{>{\columncolor{mymutedblue}}c}{\textbf{75.13}} &
\multicolumn{1}{>{\columncolor{mymutedblue}}c}{\textbf{70.73}} &
\multicolumn{1}{>{\columncolor{mymutedblue}}c}{\textbf{5.86}} &
\multicolumn{1}{>{\columncolor{mymutedblue}}c}{\textbf{85.13}} \\
\bottomrule
\end{tabular}%
}
\end{table}

\begin{table}[t]
\centering
\caption{Main results on visual question answering.}
\label{tab:vqa}
\renewcommand{\arraystretch}{0.96}
\resizebox{0.98\columnwidth}{!}{%
\begin{tabular}{lcccc}
\toprule
Method & Clean Acc. & Pert Acc. & RPD $\downarrow$ & CCA $\uparrow$ \\
\midrule
\rowcolor{gray!30}
\multicolumn{5}{l}{\textbf{\textit{General-domain}}} \\
LLaVA-1.5-7B~\cite{liu2024improved} & 62.32 & 48.69 & 21.87 & \underline{87.74} \\
Qwen3-VL-8B-Instruct~\cite{bai2025qwen3} & 71.38 & 58.34 & 18.26 & 71.31 \\
\rowcolor{gray!30}
\multicolumn{5}{l}{\textbf{\textit{RS-Specialized}}} \\
GeoChat~\cite{kuckreja2024geochat} & 80.98 & 48.97 & 39.53 & 72.64 \\
GeoPix~\cite{ou2025geopix} & 62.76 & 54.47 & \underline{13.21} & 65.91 \\
RemoteReasoner~\cite{yao2025remotereasoner} & 68.68 & 53.37 & 22.30 & 66.37 \\
EarthDial~\cite{soni2025earthdial} & \textbf{92.84} & \underline{79.96} & 13.87 & 82.87 \\
\rowcolor{mymutedblue}
\multicolumn{1}{>{\columncolor{mymutedblue}}l}{\textbf{RemoteShield}} &
\multicolumn{1}{>{\columncolor{mymutedblue}}c}{\underline{89.47}} &
\multicolumn{1}{>{\columncolor{mymutedblue}}c}{\textbf{86.65}} &
\multicolumn{1}{>{\columncolor{mymutedblue}}c}{\textbf{3.15}} &
\multicolumn{1}{>{\columncolor{mymutedblue}}c}{\textbf{91.72}} \\
\bottomrule
\end{tabular}%
}
\end{table}

\subsection{Training Data}

Starting from the semantically equivalent condition set $\mathcal{X}_i$, we construct the preference supervision used for DPO training. The core idea is to compare responses generated under matched perturbation conditions and use their quality contrast with respect to the shared reference target $y_i^\star$ to identify preferred and rejected responses.

\subsubsection{Multiple Response Inference}

Starting from the initialization policy $\pi_{\mathrm{base}}$ initialized from Qwen3-VL-8B-Instruct, we draw $N$ stochastic responses under each condition $x_i^{(j)} \in \mathcal{X}_i$:

\begin{equation}
\resizebox{0.9\linewidth}{!}{%
$\displaystyle
o_{i,j}^{(n)} \sim \pi_{\mathrm{base}}(\cdot \mid x_i^{(j)}),
\qquad
j \in \{1,\dots,4\},\;
n \in \{1,\dots,N\}.
$%
}
\label{eq:sc_sampling}
\end{equation}
Collecting responses from all four conditions yields the candidate pool $\mathcal{O}_i=\{o_{i,j}^{(n)}\}$ for subsequent scoring and preference selection.

\begin{table}[t]
\centering
\caption{Main results on visual grounding.}
\label{tab:visual_grounding}
\renewcommand{\arraystretch}{1.15}
\resizebox{\columnwidth}{!}{%
\begin{tabular}{lcccccc}
\toprule
\multirow{2}{*}{Method} & \multicolumn{2}{c}{Clean} & \multicolumn{2}{c}{Perturbed} & \multirow{2}{*}{RPD $\downarrow$} & \multirow{2}{*}{CCA $\uparrow$} \\
\cmidrule(lr){2-3} \cmidrule(lr){4-5}
& Acc@0.5 & gIoU & Acc@0.5 & gIoU & & \\
\midrule
\rowcolor{gray!30}
\multicolumn{7}{l}{\textbf{\textit{General-domain}}} \\
InternVL3.5-8B~\cite{wang2025internvl3} & \underline{55.20} & \underline{53.30} & 45.55 & \underline{46.42} & 17.48 & 57.54 \\
Qwen3-VL-8B-Instruct~\cite{bai2025qwen3} & 53.95 & 47.46 & \underline{49.35} & 43.18 & \underline{8.53} & \underline{59.34} \\
\rowcolor{gray!30}
\multicolumn{7}{l}{\textbf{\textit{RS-Specialized}}} \\
GeoChat~\cite{kuckreja2024geochat} & 25.15 & 29.69 & 7.35 & 10.00 & 70.78 & 7.74 \\
GeoPix~\cite{ou2025geopix} & 33.30 & 33.34 & 18.20 & 20.92 & 45.35 & 24.58 \\
RemoteReasoner~\cite{yao2025remotereasoner} & 36.85 & 37.73 & 27.75 & 29.44 & 24.69 & 41.56 \\
EarthDial~\cite{soni2025earthdial} & 31.65 & 32.38 & 27.60 & 29.15 & 12.80 & 27.63 \\
\rowcolor{mymutedblue}
\multicolumn{1}{>{\columncolor{mymutedblue}}l}{RemoteShield} &
\multicolumn{1}{>{\columncolor{mymutedblue}}c}{\textbf{68.50}} &
\multicolumn{1}{>{\columncolor{mymutedblue}}c}{\textbf{57.03}} &
\multicolumn{1}{>{\columncolor{mymutedblue}}c}{\textbf{64.40}} &
\multicolumn{1}{>{\columncolor{mymutedblue}}c}{\textbf{53.10}} &
\multicolumn{1}{>{\columncolor{mymutedblue}}c}{\textbf{5.99}} &
\multicolumn{1}{>{\columncolor{mymutedblue}}c}{\textbf{65.86}} \\
\bottomrule
\end{tabular}%
}
\end{table}

\begin{table}[t]
\centering
\caption{Overall framework ablation on visual grounding.}
\label{tab:sft_visual_grounding}
\renewcommand{\arraystretch}{1.15}
\resizebox{0.98\columnwidth}{!}{%
\begin{tabular}{lcccccc}
\toprule
\multirow{2}{*}{Method} & \multicolumn{2}{c}{Clean} & \multicolumn{2}{c}{Perturbed} & \multirow{2}{*}{RPD $\downarrow$} & \multirow{2}{*}{CCA $\uparrow$} \\
\cmidrule(lr){2-3} \cmidrule(lr){4-5}
& Acc@0.5 & gIoU & Acc@0.5 & gIoU & & \\
\midrule
Base Model & 53.95 & 47.46 & \underline{49.35} & 43.18 & 8.53 & \underline{59.34} \\
Clean-SFT & \underline{56.40} & \underline{49.30} & 46.25 & 41.66 & 18.00 & 46.05 \\
Mix-SFT & 51.05 & 46.83 & 48.20 & \underline{44.56} & \textbf{5.58} & 52.30 \\
\rowcolor{mymutedblue}
\multicolumn{1}{>{\columncolor{mymutedblue}}l}{RemoteShield} &
\multicolumn{1}{>{\columncolor{mymutedblue}}c}{\textbf{68.50}} &
\multicolumn{1}{>{\columncolor{mymutedblue}}c}{\textbf{57.03}} &
\multicolumn{1}{>{\columncolor{mymutedblue}}c}{\textbf{64.40}} &
\multicolumn{1}{>{\columncolor{mymutedblue}}c}{\textbf{53.10}} &
\multicolumn{1}{>{\columncolor{mymutedblue}}c}{\underline{5.99}} &
\multicolumn{1}{>{\columncolor{mymutedblue}}c}{\textbf{65.86}} \\
\bottomrule
\end{tabular}%
}
\end{table}

\subsubsection{Unified Quality Scoring}

Given the candidate pool $\mathcal{O}_i$, we assign each sampled response a unified quality score
\begin{equation}
s_i(o)=\mathcal{S}(o,y_i^\star),
\qquad
s_i(o)\in[0,1],
\label{eq:sc_score}
\end{equation}
where $y_i^\star$ denotes the shared reference target for the semantically equivalent condition set $\mathcal{X}_i$. The scoring function $\mathcal{S}$ is instantiated according to answer structure: exact matching for discrete-valued answers, relative numerical error for count-valued answers, and Hungarian-matched average IoU for coordinate-valued answers.

For \textbf{discrete-valued answers}, including scene classification, yes/no VQA, and short-answer VQA with finite response spaces, we use normalized exact matching:
\begin{equation}
\mathcal{S}_{\mathrm{dis}}(o,y_i^\star)=
\begin{cases}
1, & \operatorname{norm}(o)=\operatorname{norm}(y_i^\star),\\
0, & \text{otherwise},
\end{cases}
\label{eq:sc_discrete_score}
\end{equation}
where $\operatorname{norm}(\cdot)$ denotes standard text normalization.

For \textbf{count-valued answers}, we measure relative numerical error so that near-correct predictions remain clearly distinguishable from severely incorrect ones. Let $p=\operatorname{cnt}(o)$ and $g=\operatorname{cnt}(y_i^\star)$ denote the predicted and reference counts extracted from the generated response and target, respectively. We define
\begin{equation}
\resizebox{0.9\linewidth}{!}{%
$\displaystyle
\mathcal{S}_{\mathrm{cnt}}(o,y_i^\star)=
\begin{cases}
1, & p=g,\\[3pt]
0, & (g=0 \land p\neq g)\ \text{or}\ \dfrac{|p-g|}{|g|}>0.5,\\[8pt]
\exp\!\left(-3\dfrac{|p-g|}{|g|}\right), & \text{otherwise}.
\end{cases}
$%
}
\label{eq:sc_counting_score}
\end{equation}

For \textbf{coordinate-valued answers}, as in visual grounding, we parse the prediction $o$ and the reference target $y_i^\star$ into sets of axis-aligned bounding boxes $P$ and $G$, respectively, and score them by the Hungarian-matched average IoU between box sets:
\begin{equation}
\mathcal{S}_{\mathrm{grd}}(o,y_i^\star)=
\frac{1}{|G|}
\sum_{(b_g,b_p)\in\operatorname{match}(G,P)}
\operatorname{IoU}(b_g,b_p).
\label{eq:sc_grounding_score}
\end{equation}

\begin{table}[t]
\centering
\caption{Preference data ablation on visual grounding.}
\label{tab:ablation_self_consistency}
\resizebox{\columnwidth}{!}{%
\begin{tabular}{lcccccc}
\toprule
\multirow{2}{*}{Variant} & \multicolumn{2}{c}{Clean} & \multicolumn{2}{c}{Perturbed} & \multirow{2}{*}{RPD $\downarrow$} & \multirow{2}{*}{CCA $\uparrow$} \\
\cmidrule(lr){2-3}\cmidrule(lr){4-5}
& Acc@0.5 & gIoU & Acc@0.5 & gIoU & & \\
\midrule
Base Model & 53.95 & 47.46 & 49.35 & 43.18 & \underline{8.53} & 59.34 \\
w/o multi-sampling & \underline{67.00} & \underline{56.18} & \underline{60.65} & \underline{51.09} & 9.48 & 61.83 \\
Clean-only generation & 63.55 & 53.36 & 56.75 & 47.62 & 10.70 & \underline{64.26} \\
Pert-only generation & 64.65 & 53.27 & 57.10 & 48.11 & 10.29 & 62.79 \\
\rowcolor{mymutedblue}
RemoteShield & \textbf{68.50} & \textbf{57.03} & \textbf{64.40} & \textbf{53.10} & \textbf{5.99} & \textbf{65.86} \\
\bottomrule
\end{tabular}%
}
\end{table}

\subsubsection{Preference Pair Selection}

Given the scored candidate pool $\mathcal{O}_i$, we define the preferred and rejected responses as
\begin{equation}
y_i^{w}=\arg\max_{o\in\mathcal{O}_i}s_i(o),
\qquad
y_i^{l}=\arg\min_{o\in\mathcal{O}_i}s_i(o),
\label{eq:pref_selection}
\end{equation}
where $y_i^{w}$ and $y_i^{l}$ denote the highest-scoring and lowest-scoring responses, respectively. Since all candidates in $\mathcal{O}_i$ are sampled from the semantically equivalent condition set $\mathcal{X}_i$, this yields a cluster-level preference pair.

To form DPO training data, we instantiate this pair on the clean condition $x_i^{(1)}$ and the jointly perturbed condition $x_i^{(4)}$, yielding
\begin{equation}
\mathcal{D}_{\mathrm{pref}}
=
\bigcup_i
\left\{
\bigl(x_i^{(1)},y_i^{w},y_i^{l}\bigr),
\bigl(x_i^{(4)},y_i^{w},y_i^{l}\bigr)
\right\}.
\label{eq:pref_dataset}
\end{equation}
The intermediate conditions $x_i^{(2)}$ and $x_i^{(3)}$ are used only during response inference and scoring.

\subsection{Direct Preference Optimization}

Given the preference corpus $\mathcal{D}_{\mathrm{pref}}$, the remaining step is to align the policy with the selected preferred--rejected response pairs. To this end, we adopt standard Direct Preference Optimization (DPO)~\cite{rafailov2023direct}, which optimizes the policy directly from pairwise preferences without introducing a separately trained reward model. Each training instance is a triplet $(x_i,y_i^w,y_i^l)$, where $x_i$ denotes either the clean condition $x_i^{(1)}$ or the jointly perturbed condition $x_i^{(4)}$, and $y_i^w,y_i^l$ are the corresponding preferred and rejected responses.

Let $\pi_{\mathrm{ref}}$ denote a frozen copy of $\pi_{\mathrm{base}}$, and let $\beta>0$ control regularization strength. We first define the preference logit
\begin{equation}
\Delta_i
=
\beta
\left(
\log \frac{\pi_\theta(y_i^w \mid x_i)}{\pi_{\mathrm{ref}}(y_i^w \mid x_i)}
-
\log \frac{\pi_\theta(y_i^l \mid x_i)}{\pi_{\mathrm{ref}}(y_i^l \mid x_i)}
\right),
\label{eq:dpo_logit}
\end{equation}
and optimize the policy with the standard DPO objective
\begin{equation}
\mathcal{L}_{\mathrm{DPO}}
=
-
\mathbb{E}_{(x_i,y_i^w,y_i^l)\sim\mathcal{D}_{\mathrm{pref}}}
\left[
\log \sigma(\Delta_i)
\right].
\label{eq:dpo_loss}
\end{equation}
This objective increases the relative likelihood of the preferred response over the rejected one while keeping the learned policy close to the reference policy. Because $\mathcal{D}_{\mathrm{pref}}$ contains triplets instantiated from both clean and jointly perturbed conditions of the same task instance, optimizing Eq.~\eqref{eq:dpo_loss} encourages the policy to remain consistent across semantically equivalent multimodal inputs.

\begin{table}[t]
\centering
\caption{Training method ablation on visual grounding.}
\label{tab:ablation_preference_optimization}
\renewcommand{\arraystretch}{1.05}
\resizebox{0.98\columnwidth}{!}{%
\begin{tabular}{lcccccc}
\toprule
\multirow{2}{*}{Variant} & \multicolumn{2}{c}{Clean} & \multicolumn{2}{c}{Perturbed} & \multirow{2}{*}{RPD $\downarrow$} & \multirow{2}{*}{CCA $\uparrow$} \\
\cmidrule(lr){2-3}\cmidrule(lr){4-5}
& Acc@0.5 & gIoU & Acc@0.5 & gIoU & & \\
\midrule
Base Model & 53.95 & 47.46 & 49.35 & 43.18 & 8.53 & 59.34 \\
Preferred-SFT & \underline{62.15} & \underline{52.44} & \underline{58.20} & \underline{49.37} & \underline{6.36} & \underline{64.14} \\
Two-turn DPO & 54.30 & 47.74 & 48.40 & 41.93 & 10.87 & 58.13 \\
\rowcolor{mymutedblue}
RemoteShield & \textbf{68.50} & \textbf{57.03} & \textbf{64.40} & \textbf{53.10} & \textbf{5.99} & \textbf{65.86} \\
\bottomrule
\end{tabular}%
}
\end{table}
\section{Experiments}

To evaluate whether RemoteShield improves robustness and behavioral stability under semantically equivalent multimodal perturbations, we conduct experiments on the clean--perturbed benchmarks under the evaluation protocol established in Sec.~\ref{sec:robustness_evaluation}. We present the main comparisons with representative baselines, followed by ablation studies and further robustness analyses.

\begin{figure*}[!t]
    \centering
    \includegraphics[width=\textwidth]{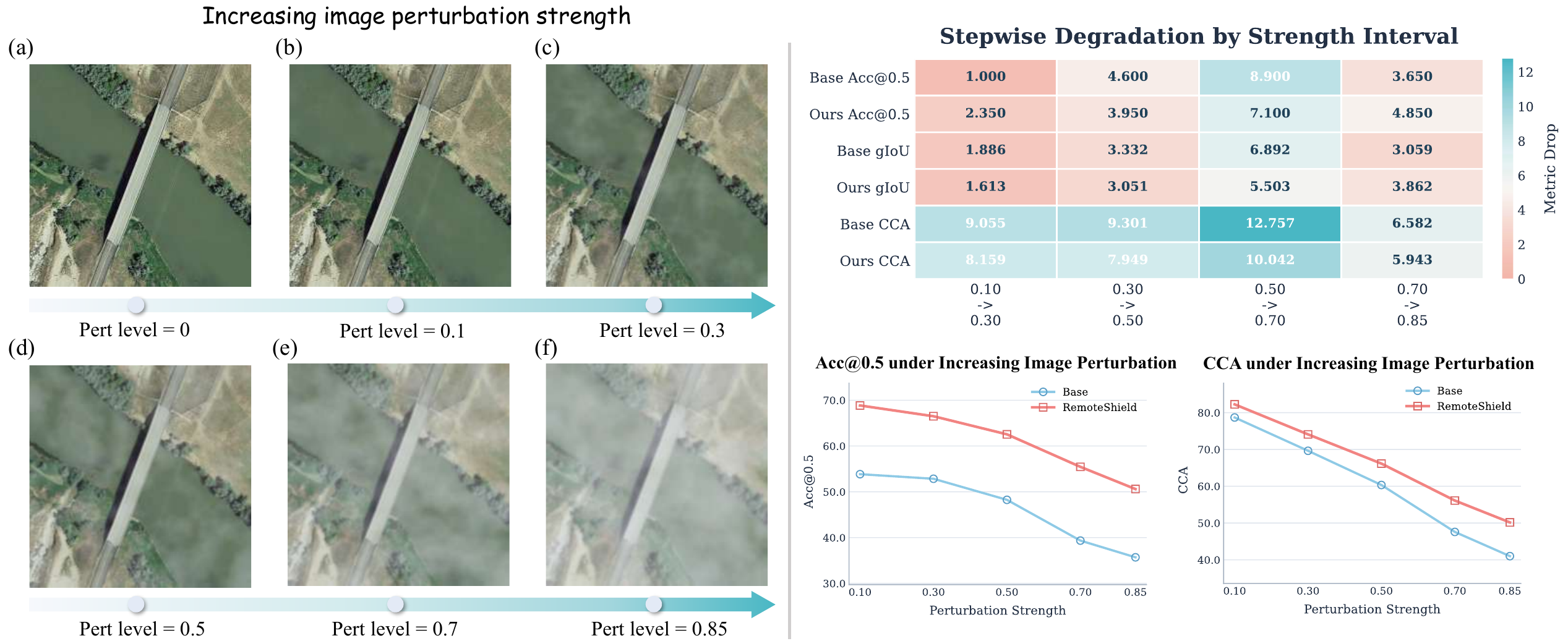}
    \caption{Performance under increasing image perturbation strength.}
    \label{fig:Increasing image perturbation strength}
\end{figure*}

\subsection{Implementation Details}

RemoteShield is trained with Direct Preference Optimization (DPO) using the ms-swift~\cite{zhao2025swift} framework and DeepSpeed ZeRO-2~\cite{rasley2020deepspeed}. We initialize from Qwen3-VL-8B-Instruct~\cite{bai2025qwen3} and apply LoRA~\cite{hu2022lora} ($r=32, \alpha=64$) to all linear layers. Training is conducted in bfloat16 precision with SDPA attention. The model is trained for 8 epochs with a learning rate of $3\times 10^{-5}$ and a warmup ratio of $0.05$. For DPO, we set $\beta=0.1$, use the sigmoid loss, and set $\texttt{rpo\_alpha}=0.1$. The per-device batch size is 1 with gradient accumulation over 16 steps, yielding an effective batch size of 96 on 6 NVIDIA H100 GPUs.

\subsection{Main Results}

We now report the main comparison on scene classification, VQA, and visual grounding in Tables~\ref{tab:scene_classification},~\ref{tab:vqa}, and~\ref{tab:visual_grounding}, respectively. Because RemoteShield is designed to improve robustness and cross-condition consistency under multimodal perturbations, we center the analysis on RPD and CCA, using standard task metrics as supporting evidence. Across all three tasks, RemoteShield consistently achieves the lowest RPD and the highest CCA, indicating the smallest perturbation-induced degradation and the strongest behavioral stability among all compared models.

\subsubsection{Scene Classification}

Table~\ref{tab:scene_classification} reports the results on scene classification. RemoteShield achieves the best robustness profile, with an RPD of 5.86 and a CCA of 85.13. Compared with the strongest general-domain baseline, Qwen3-VL-8B-Instruct, it reduces RPD from 7.18 to 5.86 and improves CCA from 82.27 to 85.13. The margin is larger against RS-specialized models: relative to RemoteReasoner, RemoteShield lowers RPD from 16.42 to 5.86 and raises CCA from 68.10 to 85.13. This gain is accompanied by the strongest perturbed accuracy of 70.73, showing that improved robustness does not come at the expense of task performance.

\subsubsection{Visual Question Answering}

Table~\ref{tab:vqa} shows an even larger advantage on VQA. Among the baselines, the best RPD is 13.21 and the best CCA is 87.74, while RemoteShield reaches 3.15 RPD and 91.72 CCA. Its perturbed accuracy also improves from the strongest baseline value of 79.96 to 86.65. These results indicate that RemoteShield not only preserves answer correctness under perturbation, but also maintains stronger consistency across matched clean and perturbed conditions.

\subsubsection{Visual Grounding}

The same pattern holds for visual grounding in Table~\ref{tab:visual_grounding}, where perturbations translate more directly into localization drift. RemoteShield achieves the lowest RPD of 5.99 and the highest CCA of 65.86, improving over the strongest baseline values of 8.53 and 59.34. The improvement is also reflected in perturbed grounding performance, where Acc@0.5 rises from 49.35 to 64.40 and gIoU rises from 46.42 to 53.10. This indicates that the advantage of RemoteShield lies not only in stronger localization accuracy, but also in better resistance to perturbation and more stable spatial behavior across semantically equivalent conditions.

Overall, RemoteShield consistently improves the two target properties: robustness to perturbation and cross-condition consistency. Gains in perturbed task metrics show that improved robustness is also reflected in stronger task performance under perturbation.

\subsection{Ablation Studies}

Having established the robustness advantage of RemoteShield, we next examine which parts of the training framework are responsible for it. We answer this question through ablations on visual grounding, where perturbation effects appear clearly as localization drift. Full variant definitions are deferred to the supplementary material.

\subsubsection{Ablation on Overall Framework}

Using the same backbone, we compare the base model, Clean-SFT, Mix-SFT, and RemoteShield (Table~\ref{tab:sft_visual_grounding}). Clean-SFT improves clean grounding performance but sharply weakens robustness, pushing RPD from 8.53 to 18.00 and lowering CCA from 59.34 to 46.05. Mix-SFT seems to move in a more favorable direction, but its smaller RPD of 5.58 is misleading: perturbed Acc@0.5 drops below the base model (48.20 vs.~49.35) and CCA remains low at 52.30, indicating weaker predictions on both clean and perturbed inputs rather than greater robustness. Only RemoteShield improves perturbed performance while maintaining a strong robustness profile, with perturbed Acc@0.5 of 64.40, RPD of 5.99, and CCA of 65.86. This shows that robustness is not recovered by exposure to perturbed samples alone; the main gain comes from preference-based alignment across semantically equivalent conditions.

\begin{figure}[t]
    \centering
    \includegraphics[width=\columnwidth]{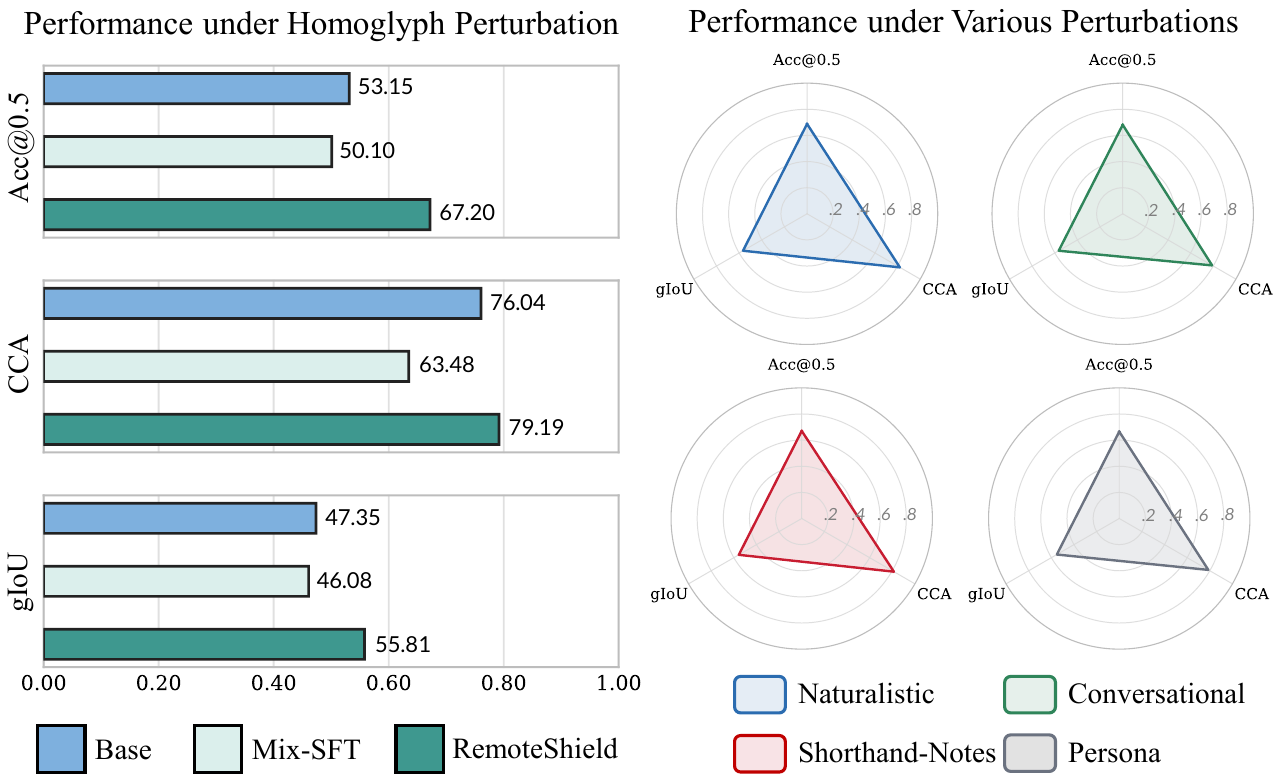}
    \caption{RemoteShield outperforms baselines under four seen text perturbations and an unseen homoglyph perturbation, showing robustness beyond training perturbations.}
    \label{fig:text_pertur}
\end{figure}

\subsubsection{Ablation on Preference Data}

To study how preference data are prepared before DPO, we compare the full design with three variants in Table~\ref{tab:ablation_self_consistency}. These include \textit{w/o multi-sampling}, which uses one response per condition, together with \textit{Clean-only generation} and \textit{Pert-only generation}, which build candidate pools from clean and perturbed conditions, respectively. Without multi-sampling, RPD rises from 5.99 to 9.48 and CCA falls from 65.86 to 61.83, with perturbed Acc@0.5 dropping from 64.40 to 60.65. Restricting candidate generation to one side of the clean--perturbed contrast is also suboptimal: both Clean-only and Pert-only generation raise RPD above 10 and remain below the full design in consistency and perturbed performance. The benefit of this stage lies in preference quality: reliable preferred--rejected pairs require both response diversity and joint coverage of clean and perturbed conditions.

\subsubsection{Ablation on Training Methods}

To study how the constructed preference pairs are used during training, we compare RemoteShield with two alternatives: \textit{Preferred-SFT}, which trains only on the preferred response; and \textit{Two-turn DPO}, which rewrites the clean and perturbed inputs as a two-turn dialogue while keeping the DPO objective (Table~\ref{tab:ablation_preference_optimization}). Preferred-SFT improves over the base model, lowering RPD to 6.36 and raising CCA to 64.14, but it still trails RemoteShield and falls short on perturbed Acc@0.5 (58.20 vs.~64.40). This gap shows that preferred responses alone are not enough: rejected responses provide the contrastive signal that steers the model away from unstable outputs. Two-turn DPO performs worse than the base model itself, with an RPD of 10.87 and a CCA of 58.13, suggesting that clean and perturbed inputs are better treated as parallel variants of the same instance than as sequential turns. Overall, preference learning is most effective under the single-turn formulation used in RemoteShield.

\begin{figure}[!t]
    \centering
    \includegraphics[width=\columnwidth]{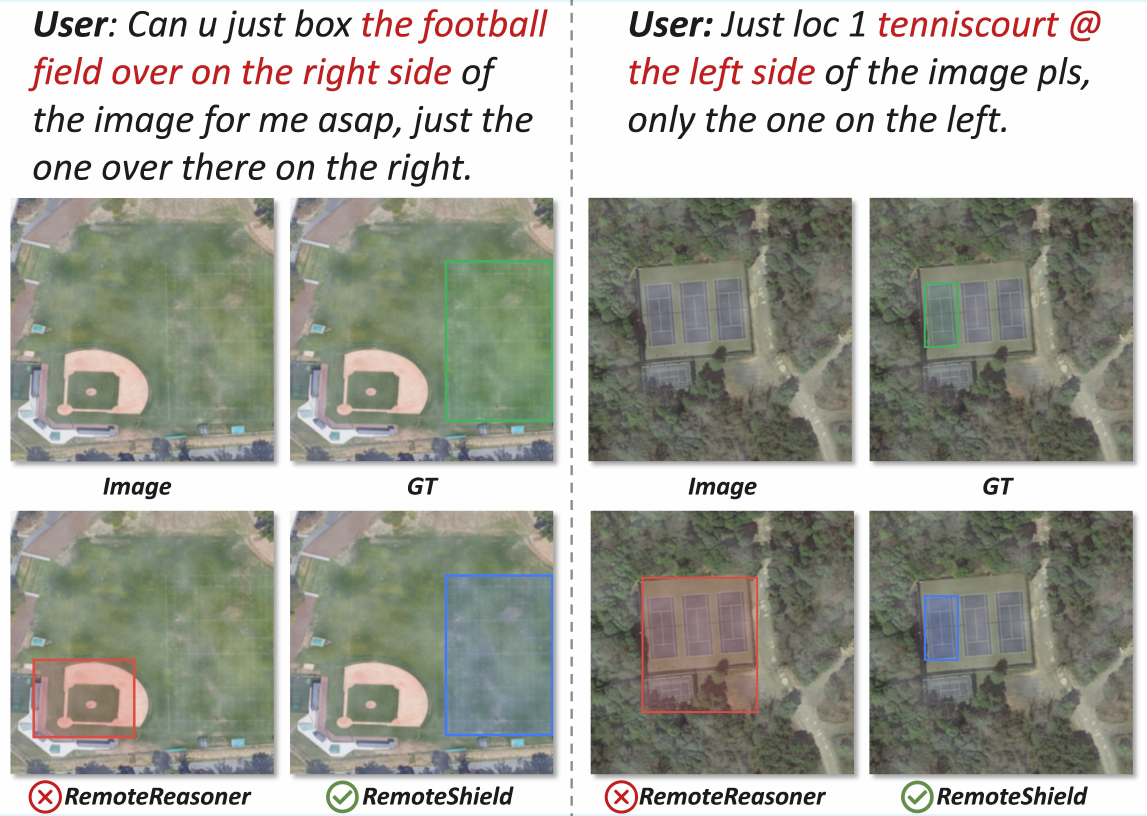}
    \caption{Qualitative case studies under perturbation.}
   
    \label{fig:case_study}
\end{figure}

\subsection{Further Analysis}

The main results show that RemoteShield is robust under the perturbation setting, while the ablations clarify where this gain comes from. We further examine whether this advantage holds under stronger image corruption, whether it transfers to unseen text perturbations, and how it appears in qualitative cases.

\subsubsection{Stability under Stronger Image Perturbations}

The main comparison is conducted at a fixed perturbation level, but a stronger test of robustness is whether the gain of RemoteShield persists as image corruption intensifies over a broad range of perturbation strengths. Fig.~\ref{fig:Increasing image perturbation strength} shows that increasing the perturbation level from 0.10 to 0.85 reduces the base model's Acc@0.5 from 53.85 to 35.70 and its CCA from 78.70 to 41.00, whereas RemoteShield declines more gradually, with Acc@0.5 decreasing from 68.85 to 50.60 and CCA from 82.27 to 50.17. The contrast is clearest in the 0.50--0.70 interval, where the base model loses 8.90 in Acc@0.5, 6.89 in gIoU, and 12.76 in CCA, compared with drops of 7.10, 5.50, and 10.04 for RemoteShield. The gain therefore lies not only in stronger performance at individual perturbation levels, but also in a flatter degradation trajectory as perturbation severity increases.

\subsubsection{Generalization to Unseen Text Perturbations}

Robustness on seen perturbation types does not necessarily imply robustness to unseen ones. As shown in Fig.~\ref{fig:text_pertur}, we therefore introduce an unseen homoglyph perturbation, in which some English characters are replaced by visually identical Cyrillic characters. Under this unseen corruption, RemoteShield still outperforms both the base model and Mix-SFT on all three metrics, achieving 67.20 Acc@0.5, 55.81 gIoU, and 79.19 CCA. This indicates that the robustness learned by RemoteShield transfers beyond the perturbation patterns observed during training.

\subsubsection{Qualitative Robustness Analysis}

Fig.~\ref{fig:case_study} reveals a consistent difference in failure mode under perturbation. RemoteReasoner tends to drift toward semantically related but incorrect regions, or to predict overly broad boxes, whereas RemoteShield remains closer to the ground truth. In the left example, the query refers to the football field on the right side of the image, yet RemoteReasoner localizes the baseball field. In the right example, the query asks for only the tennis court on the left, but RemoteReasoner predicts a broader region. In both cases, RemoteShield preserves the intended semantic and spatial constraints, consistent with the quantitative results above.
\section{Conclusion}

In this work, we identify the robustness gap of remote sensing MLLMs under realistic multimodal perturbations as a fundamental obstacle to practical Earth Observation. We propose RemoteShield, a robust remote sensing MLLM that learns stable outputs across semantically equivalent clean and perturbed conditions through cross-condition preference learning. Experiments on scene classification, VQA and visual grounding show that RemoteShield consistently improves robustness and behavioral stability over both general-domain and RS-specific baselines, moving EO MLLMs a step closer to reliable real-world deployment.

{
    \small
    \bibliographystyle{ieeenat_fullname}
    \bibliography{main}

@String(CVPR= {IEEE Conf. Comput. Vis. Pattern Recog.})

@String(ICLR = {Int. Conf. Learn. Represent.})

@String(AAAI = {AAAI})

@String(CVPR  = {CVPR})

@String(ICLR  = {ICLR})

@inproceedings{remotesam,
  title={Remotesam: Towards segment anything for earth observation},
  author={Yao, Liang and Liu, Fan and Chen, Delong and Zhang, Chuanyi and Wang, Yijun and Chen, Ziyun and Xu, Wei and Di, Shimin and Zheng, Yuhui},
  booktitle={Proceedings of the 33rd ACM International Conference on Multimedia},
  year={2025}
}

@inproceedings{directsam_rs,
  title={Prompting directsam for semantic contour extraction in remote sensing images},
  author={Miao, Shiyu and Chen, Delong and Liu, Fan and Zhang, Chuanyi and Gu, Yanhui and Guo, Shengjie and Zhou, Jun},
  booktitle={2025-2025 IEEE International Conference on Acoustics, Speech and Signal Processing},
  year={2025},
  organization={IEEE}
}

@article{reobench,
  title={REOBench: Benchmarking Robustness of Earth Observation Foundation Models},
  author={Li, Xiang and Tao, Yong and Zhang, Siyuan and Liu, Siwei and Xiong, Zhitong and Luo, Chunbo and Liu, Lu and Pechenizkiy, Mykola and Zhu, Xiao Xiang and Huang, Tianjin},
  journal={arXiv preprint arXiv:2505.16793},
  year={2025}
}

@inproceedings{wang2024earthvqa,
  title={Earthvqa: Towards queryable earth via relational reasoning-based remote sensing visual question answering},
  author={Wang, Junjue and Zheng, Zhuo and Chen, Zihang and Ma, Ailong and Zhong, Yanfei},
  booktitle={AAAI},
  volume={38},
  number={6},
  pages={5481--5489},
  year={2024}
}

@article{yao2025falcon,
  title={Falcon: A Remote Sensing Vision-Language Foundation Model},
  author={Yao, Kelu and Xu, Nuo and Yang, Rong and Xu, Yingying and Gao, Zhuoyan and Kitrungrotsakul, Titinunt and Ren, Yi and Zhang, Pu and Wang, Jin and Wei, Ning and others},
  journal={arXiv preprint arXiv:2503.11070},
  year={2025}
}

@inproceedings{rasley2020deepspeed,
  title={Deepspeed: System optimizations enable training deep learning models with over 100 billion parameters},
  author={Rasley, Jeff and Rajbhandari, Samyam and Ruwase, Olatunji and He, Yuxiong},
  booktitle={Proceedings of the 26th ACM SIGKDD international conference on knowledge discovery \& data mining},
  pages={3505--3506},
  year={2020}
}

@article{hu2022lora,
  title={Lora: Low-rank adaptation of large language models.},
  author={Hu, Edward J and Shen, Yelong and Wallis, Phillip and Allen-Zhu, Zeyuan and Li, Yuanzhi and Wang, Shean and Wang, Lu and Chen, Weizhu and others},
  journal={ICLR},
  volume={1},
  number={2},
  pages={3},
  year={2022}
}

@inproceedings{kuckreja2024geochat,
  title={Geochat: Grounded large vision-language model for remote sensing},
  author={Kuckreja, Kartik and Danish, Muhammad Sohail and Naseer, Muzammal and Das, Abhijit and Khan, Salman and Khan, Fahad Shahbaz},
  booktitle={CVPR},
  pages={27831--27840},
  year={2024}
}

@article{hu2025rsgpt,
  title={Rsgpt: A remote sensing vision language model and benchmark},
  author={Hu, Yuan and Yuan, Jianlong and Wen, Congcong and Lu, Xiaonan and Liu, Yu and Li, Xiang},
  journal={ISPRS Journal of Photogrammetry and Remote Sensing},
  volume={224},
  pages={272--286},
  year={2025},
  publisher={Elsevier}
}

@article{yao2025remotereasoner,
  title={Remotereasoner: Towards unifying geospatial reasoning workflow},
  author={Yao, Liang and Liu, Fan and Lu, Hongbo and Zhang, Chuanyi and Min, Rui and Xu, Shengxiang and Di, Shimin and Peng, Pai},
  journal={arXiv preprint arXiv:2507.19280},
  year={2025}
}

@article{luo2024skysensegpt,
  title={Skysensegpt: A fine-grained instruction tuning dataset and model for remote sensing vision-language understanding},
  author={Luo, Junwei and Pang, Zhen and Zhang, Yongjun and Wang, Tingzhu and Wang, Linlin and Dang, Bo and Lao, Jiangwei and Wang, Jian and Chen, Jingdong and Tan, Yihua and others},
  journal={arXiv preprint arXiv:2406.10100},
  year={2024}
}

@article{wang2024ringmogpt,
  title={Ringmogpt: A unified remote sensing foundation model for vision, language, and grounded tasks},
  author={Wang, Peijin and Hu, Huiyang and Tong, Boyuan and Zhang, Ziqi and Yao, Fanglong and Feng, Yingchao and Zhu, Zining and Chang, Hao and Diao, Wenhui and Ye, Qixiang and others},
  journal={IEEE Transactions on Geoscience and Remote Sensing},
  volume={63},
  pages={1--20},
  year={2024},
  publisher={IEEE}
}

@article{ou2025geopix,
  title={GeoPix: A multimodal large language model for pixel-level image understanding in remote sensing},
  author={Ou, Ruizhe and Hu, Yuan and Zhang, Fan and Chen, Jiaxin and Liu, Yu},
  journal={IEEE Geoscience and Remote Sensing Magazine},
  year={2025},
  publisher={IEEE}
}

@inproceedings{soni2025earthdial,
  title={Earthdial: Turning multi-sensory earth observations to interactive dialogues},
  author={Soni, Sagar and Dudhane, Akshay and Debary, Hiyam and Fiaz, Mustansar and Munir, Muhammad Akhtar and Danish, Muhammad Sohail and Fraccaro, Paolo and Watson, Campbell D and Klein, Levente J and Khan, Fahad Shahbaz and others},
  booktitle={Proceedings of the Computer Vision and Pattern Recognition Conference},
  pages={14303--14313},
  year={2025}
}

@article{chen2023benchmarking,
  title={Benchmarking robustness of adaptation methods on pre-trained vision-language models},
  author={Chen, Shuo and Gu, Jindong and Han, Zhen and Ma, Yunpu and Torr, Philip and Tresp, Volker},
  journal={Advances in Neural Information Processing Systems},
  volume={36},
  pages={51758--51777},
  year={2023}
}

@article{jiang2025survey,
  title={Survey of adversarial robustness in multimodal large language models},
  author={Jiang, Chengze and Wang, Zhuangzhuang and Dong, Minjing and Gui, Jie},
  journal={arXiv preprint arXiv:2503.13962},
  year={2025}
}

@article{zhao2023evaluating,
  title={On evaluating adversarial robustness of large vision-language models},
  author={Zhao, Yunqing and Pang, Tianyu and Du, Chao and Yang, Xiao and Li, Chongxuan and Cheung, Ngai-Man Man and Lin, Min},
  journal={Advances in Neural Information Processing Systems},
  volume={36},
  pages={54111--54138},
  year={2023}
}

@inproceedings{qiu2025benchmarking,
  title={Benchmarking Multimodal Large Language Models Against Image Corruptions},
  author={Qiu, Xinkuan and Kan, Meina and Zhou, Yongbin and Shan, Shiguang},
  booktitle={Proceedings of the IEEE/CVF International Conference on Computer Vision},
  pages={9014--9023},
  year={2025}
}

@article{rabanser2026towards,
  title={Towards a science of AI agent reliability},
  author={Rabanser, Stephan and Kapoor, Sayash and Kirgis, Peter and Liu, Kangheng and Utpala, Saiteja and Narayanan, Arvind},
  journal={arXiv preprint arXiv:2602.16666},
  year={2026}
}

@article{usama2025analysing,
  title={Analysing the robustness of vision-language-models to common corruptions},
  author={Usama, Muhammad and Asim, Syeda Aishah and Ali, Syed Bilal and Wasim, Syed Talal and Mansoor, Umair Bin},
  journal={arXiv preprint arXiv:2504.13690},
  year={2025}
}

@article{xu2025robustflow,
  title={Robustflow: Towards robust agentic workflow generation},
  author={Xu, Shengxiang and Zhang, Jiayi and Di, Shimin and Luo, Yuyu and Yao, Liang and Liu, Hanmo and Zhu, Jia and Liu, Fan and Zhang, Min-Ling},
  journal={arXiv preprint arXiv:2509.21834},
  year={2025}
}

@ARTICLE{robust_bench_1,
  author={He, Haodong and Ding, Jian and Xu, Bowen and Xia, Gui-Song},
  journal={IEEE Transactions on Geoscience and Remote Sensing}, 
  title={On the Robustness of Object Detection Models on Aerial Images}, 
  year={2025},
  doi={10.1109/TGRS.2024.3514741}
}

@ARTICLE{robust_bench_2,
  author={Karwowska, Kinga and Siewert, Jolanta and Sekrecka, Aleksandra},
  journal={IEEE Journal of Selected Topics in Applied Earth Observations and Remote Sensing}, 
  title={Self-Attention-Enhanced Dual-Branch Network for Cloud Detection in Panchromatic Satellite Imagery}, 
  year={2026},
  doi={10.1109/JSTARS.2025.3639193}
}

@ARTICLE{robust_bench_3,
  author={Demır, Mehmet},
  journal={IEEE Access}, 
  title={Speckle Noise Reduction in SAR Images Using Rank Residual Constraint Regularization}, 
  year={2025},
  doi={10.1109/ACCESS.2025.3628472}
}

@inproceedings{zhao2025swift,
  title={Swift: a scalable lightweight infrastructure for fine-tuning},
  author={Zhao, Yuze and Huang, Jintao and Hu, Jinghan and Wang, Xingjun and Mao, Yunlin and Zhang, Daoze and Jiang, Zeyinzi and Wu, Zhikai and Ai, Baole and Wang, Ang and others},
  booktitle={Proceedings of the AAAI Conference on Artificial Intelligence},
  volume={39},
  number={28},
  pages={29733--29735},
  year={2025}
}

@article{zhou2024towards,
  title={Towards vision-language geo-foundation model: A survey},
  author={Zhou, Yue and Zhong, Zhihang and Yang, Xue},
  journal={arXiv preprint arXiv:2406.09385},
  year={2024}
}

@article{huang2025survey,
  title={A survey on remote sensing foundation models: From vision to multimodality},
  author={Huang, Ziyue and Yan, Hongxi and Zhan, Qiqi and Yang, Shuai and Zhang, Mingming and Zhang, Chenkai and Lei, YiMing and Liu, Zeming and Liu, Qingjie and Wang, Yunhong},
  journal={arXiv preprint arXiv:2503.22081},
  year={2025}
}

@article{zhu2026foundations,
  title={On the foundations of Earth foundation models},
  author={Zhu, Xiao Xiang and Xiong, Zhitong and Wang, Yi and Stewart, Adam J and Heidler, Konrad and Wang, Yuanyuan and Yuan, Zhenghang and Dujardin, Thomas and Xu, Qingsong and Shi, Yilei},
  journal={Communications Earth \& Environment},
  year={2026},
  publisher={Nature Publishing Group UK London}
}

@article{zhou2025advances,
  title={Advances on multimodal remote sensing foundation models for Earth observation downstream tasks: A survey},
  author={Zhou, Guoqing and Lihuang, Qian and Gamba, Paolo},
  journal={Remote Sensing},
  volume={17},
  number={21},
  pages={3532},
  year={2025},
  publisher={MDPI AG}
}

@article{kansakar2016review,
  title={A review of applications of satellite earth observation data for global societal benefit and stewardship of planet earth},
  author={Kansakar, Pratistha and Hossain, Faisal},
  journal={Space Policy},
  volume={36},
  pages={46--54},
  year={2016},
  publisher={Elsevier}
}

@article{binci2026earth,
  title={Earth observation and sustainable development: A systematic literature review and content analysis about the New Space Economy},
  author={Binci, Daniele},
  journal={Environmental Innovation and Societal Transitions},
  volume={59},
  pages={101088},
  year={2026},
  publisher={Elsevier}
}

@article{connors2025earth,
  title={Earth observations for climate adaptation: tracking progress towards the Global Goal on Adaptation through satellite-derived indicators},
  author={Connors, Sarah and Schneider, Rochelle and Nalau, Johanna and Hawkins, Michelle and Ferdini, Sofia and Wang, Ying and Rast, Michael and Aunan, Kristin and Aurambout, Jean-Philippe and Dowell, Mark and others},
  journal={npj Climate and Atmospheric Science},
  volume={8},
  number={1},
  pages={359},
  year={2025},
  publisher={Nature Publishing Group UK London}
}

@article{zhang2024earthgpt,
  title={EarthGPT: A universal multimodal large language model for multisensor image comprehension in remote sensing domain},
  author={Zhang, Wei and Cai, Miaoxin and Zhang, Tong and Zhuang, Yin and Mao, Xuerui},
  journal={IEEE Transactions on Geoscience and Remote Sensing},
  volume={62},
  pages={1--20},
  year={2024},
  publisher={IEEE}
}

@article{li2024vrsbench,
  title={Vrsbench: A versatile vision-language benchmark dataset for remote sensing image understanding},
  author={Li, Xiang and Ding, Jian and Elhoseiny, Mohamed},
  journal={Advances in Neural Information Processing Systems},
  volume={37},
  pages={3229--3242},
  year={2024}
}

@article{rafailov2023direct,
  title={Direct preference optimization: Your language model is secretly a reward model},
  author={Rafailov, Rafael and Sharma, Archit and Mitchell, Eric and Manning, Christopher D and Ermon, Stefano and Finn, Chelsea},
  journal={Advances in neural information processing systems},
  volume={36},
  pages={53728--53741},
  year={2023}
}

@inproceedings{liu2024improved,
  title={Improved baselines with visual instruction tuning},
  author={Liu, Haotian and Li, Chunyuan and Li, Yuheng and Lee, Yong Jae},
  booktitle={Proceedings of the IEEE/CVF conference on computer vision and pattern recognition},
  pages={26296--26306},
  year={2024}
}

@article{bai2025qwen3,
  title={Qwen3-vl technical report},
  author={Bai, Shuai and Cai, Yuxuan and Chen, Ruizhe and Chen, Keqin and Chen, Xionghui and Cheng, Zesen and Deng, Lianghao and Ding, Wei and Gao, Chang and Ge, Chunjiang and others},
  journal={arXiv preprint arXiv:2511.21631},
  year={2025}
}

@article{wang2025internvl3,
  title={Internvl3. 5: Advancing open-source multimodal models in versatility, reasoning, and efficiency},
  author={Wang, Weiyun and Gao, Zhangwei and Gu, Lixin and Pu, Hengjun and Cui, Long and Wei, Xingguang and Liu, Zhaoyang and Jing, Linglin and Ye, Shenglong and Shao, Jie and others},
  journal={arXiv preprint arXiv:2508.18265},
  year={2025}
}

@article{xia2017aid,
  title={AID: A benchmark data set for performance evaluation of aerial scene classification},
  author={Xia, Gui-Song and Hu, Jingwen and Hu, Fan and Shi, Baoguang and Bai, Xiang and Zhong, Yanfei and Zhang, Liangpei and Lu, Xiaoqiang},
  journal={IEEE Transactions on Geoscience and Remote Sensing},
  volume={55},
  number={7},
  pages={3965--3981},
  year={2017},
  publisher={IEEE}
}

@article{lobry2020rsvqa,
  title={RSVQA: Visual question answering for remote sensing data},
  author={Lobry, Sylvain and Marcos, Diego and Murray, Jesse and Tuia, Devis},
  journal={IEEE Transactions on Geoscience and Remote Sensing},
  volume={58},
  number={12},
  pages={8555--8566},
  year={2020},
  publisher={IEEE}
}

@article{wei2021finetuned,
  title={Finetuned language models are zero-shot learners},
  author={Wei, Jason and Bosma, Maarten and Zhao, Vincent Y and Guu, Kelvin and Yu, Adams Wei and Lester, Brian and Du, Nan and Dai, Andrew M and Le, Quoc V},
  journal={arXiv preprint arXiv:2109.01652},
  year={2021}
}

@article{sanh2021multitask,
  title={Multitask prompted training enables zero-shot task generalization},
  author={Sanh, Victor and Webson, Albert and Raffel, Colin and Bach, Stephen H and Sutawika, Lintang and Alyafeai, Zaid and Chaffin, Antoine and Stiegler, Arnaud and Scao, Teven Le and Raja, Arun and others},
  journal={arXiv preprint arXiv:2110.08207},
  year={2021}
}

@article{ouyang2022training,
  title={Training language models to follow instructions with human feedback},
  author={Ouyang, Long and Wu, Jeffrey and Jiang, Xu and Almeida, Diogo and Wainwright, Carroll and Mishkin, Pamela and Zhang, Chong and Agarwal, Sandhini and Slama, Katarina and Ray, Alex and others},
  journal={Advances in neural information processing systems},
  volume={35},
  pages={27730--27744},
  year={2022}
}

@inproceedings{wang2023self,
  title={Self-instruct: Aligning language models with self-generated instructions},
  author={Wang, Yizhong and Kordi, Yeganeh and Mishra, Swaroop and Liu, Alisa and Smith, Noah A and Khashabi, Daniel and Hajishirzi, Hannaneh},
  booktitle={Proceedings of the 61st annual meeting of the association for computational linguistics (volume 1: long papers)},
  pages={13484--13508},
  year={2023}
}

@inproceedings{liang2026prompt,
  title={Prompt-Robust Vision-Language Models via Meta-Finetuning},
  author={Liang, Haohui and Huang, Runlin and Du, Yingjun and Hu, Yujia and Su, Weifeng and Snoek, Cees GM},
  booktitle={The Fourteenth International Conference on Learning Representations},
  year={2026}
}

@article{dumpala2024sugarcrepe++,
  title={Sugarcrepe++ dataset: Vision-language model sensitivity to semantic and lexical alterations},
  author={Dumpala, Sri H and Jaiswal, Aman and Sastry, Chandramouli and Milios, Evangelos and Oore, Sageev and Sajjad, Hassan},
  journal={Advances in Neural Information Processing Systems},
  volume={37},
  pages={17972--18018},
  year={2024}
}

@inproceedings{shah2025analyzing,
  title={Analyzing the Sensitivity of Vision Language Models in Visual Question Answering},
  author={Shah, Monika and Balaji, Sudarshan and Sarkhel, Somdeb and Dey, Sanorita and Venugopal, Deepak},
  booktitle={Proceedings of the Fourth Workshop on Generation, Evaluation and Metrics (GEM$^2$)},
  pages={431--438},
  year={2025}
}

@article{shen2014effective,
  title={An effective thin cloud removal procedure for visible remote sensing images},
  author={Shen, Huanfeng and Li, Huifang and Qian, Yan and Zhang, Liangpei and Yuan, Qiangqiang},
  journal={ISPRS Journal of Photogrammetry and Remote Sensing},
  volume={96},
  pages={224--235},
  year={2014},
  publisher={Elsevier}
}

@article{long2013single,
  title={Single remote sensing image dehazing},
  author={Long, Jiao and Shi, Zhenwei and Tang, Wei and Zhang, Changshui},
  journal={IEEE Geoscience and Remote Sensing Letters},
  volume={11},
  number={1},
  pages={59--63},
  year={2013},
  publisher={IEEE}
}

@inproceedings{chou2025mm,
  title={MM-R3: On (in-) consistency of vision-language models (VLMs)},
  author={Chou, Shih-Han and Chandhok, Shivam and Little, Jim and Sigal, Leonid},
  booktitle={Findings of the Association for Computational Linguistics: ACL 2025},
  pages={4762--4788},
  year={2025}
}

@article{saxena2026vlm,
  title={VLM-RobustBench: A Comprehensive Benchmark for Robustness of Vision-Language Models},
  author={Saxena, Rohit and Suglia, Alessandro and Minervini, Pasquale},
  journal={arXiv preprint arXiv:2603.06148},
  year={2026}
}

@inproceedings{alhamoud2025vision,
  title={Vision-language models do not understand negation},
  author={Alhamoud, Kumail and Alshammari, Shaden and Tian, Yonglong and Li, Guohao and Torr, Philip HS and Kim, Yoon and Ghassemi, Marzyeh},
  booktitle={Proceedings of the Computer Vision and Pattern Recognition Conference},
  pages={29612--29622},
  year={2025}
}

@article{li2025segearth,
  title={Segearth-r1: Geospatial pixel reasoning via large language model},
  author={Li, Kaiyu and Xin, Zepeng and Pang, Li and Pang, Chao and Deng, Yupeng and Yao, Jing and Xia, Guisong and Meng, Deyu and Wang, Zhi and Cao, Xiangyong},
  journal={arXiv preprint arXiv:2504.09644},
  year={2025}
}

@article{liu2025towards,
  title={Towards Faithful Reasoning in Remote Sensing: A Perceptually-Grounded GeoSpatial Chain-of-Thought for Vision-Language Models},
  author={Liu, Jiaqi and Sun, Lang and Fu, Ronghao and Yang, Bo},
  journal={arXiv preprint arXiv:2509.22221},
  year={2025}
}

@article{irvin2024teochat,
  title={Teochat: A large vision-language assistant for temporal earth observation data},
  author={Irvin, Jeremy Andrew and Liu, Emily Ruoyu and Chen, Joyce Chuyi and Dormoy, Ines and Kim, Jinyoung and Khanna, Samar and Zheng, Zhuo and Ermon, Stefano},
  journal={arXiv preprint arXiv:2410.06234},
  year={2024}
}

@article{zhang2025geo,
  title={Geo-R1: Improving Few-Shot Geospatial Referring Expression Understanding with Reinforcement Fine-Tuning},
  author={Zhang, Zilun and Guan, Zian and Zhao, Tiancheng and Shen, Haozhan and Li, Tianyu and Cai, Yuxiang and Su, Zhonggen and Liu, Zhaojun and Yin, Jianwei and Li, Xiang},
  journal={arXiv preprint arXiv:2509.21976},
  year={2025}
}

@article{wang2025geozero,
  title={GeoZero: Incentivizing Reasoning from Scratch on Geospatial Scenes},
  author={Wang, Di and Liu, Shunyu and Jiang, Wentao and Wang, Fengxiang and Liu, Yi and Qin, Xiaolei and Luo, Zhiming and Zhou, Chaoyang and Guo, Haonan and Zhang, Jing and others},
  journal={arXiv preprint arXiv:2511.22645},
  year={2025}
}

@article{jiang2025eaglevision,
  title={EagleVision: Object-level attribute multimodal LLM for remote sensing},
  author={Jiang, Hongxiang and Yin, Jihao and Wang, Qixiong and Feng, Jiaqi and Chen, Guo},
  journal={arXiv preprint arXiv:2503.23330},
  year={2025}
}

@article{xue2024reo,
  title={Reo-vlm: Transforming vlm to meet regression challenges in earth observation},
  author={Xue, Xizhe and Wei, Guoting and Chen, Hao and Zhang, Haokui and Lin, Feng and Shen, Chunhua and Zhu, Xiao Xiang},
  journal={arXiv preprint arXiv:2412.16583},
  year={2024}
}

@article{liu2024rsunivlm,
  title={Rsunivlm: A unified vision language model for remote sensing via granularity-oriented mixture of experts},
  author={Liu, Xu and Lian, Zhouhui},
  journal={arXiv preprint arXiv:2412.05679},
  year={2024}
}

@article{zhang2024earthmarker,
  title={EarthMarker: A visual prompting multimodal large language model for remote sensing},
  author={Zhang, Wei and Cai, Miaoxin and Zhang, Tong and Zhuang, Yin and Li, Jun and Mao, Xuerui},
  journal={IEEE Transactions on Geoscience and Remote Sensing},
  volume={63},
  pages={1--19},
  year={2024},
  publisher={IEEE}
}

@article{li2026georeason,
  title={GeoReason: Aligning Thinking And Answering In Remote Sensing Vision-Language Models Via Logical Consistency Reinforcement Learning},
  author={Li, Wenshuai and Xiang, Xiantai and Wen, Zixiao and Zhou, Guangyao and Niu, Ben and Wang, Feng and Huang, Lijia and Wang, Qiantong and Hu, Yuxin},
  journal={arXiv preprint arXiv:2601.04118},
  year={2026}
}

@article{fiaz2025geovlm,
  title={Geovlm-r1: Reinforcement fine-tuning for improved remote sensing reasoning},
  author={Fiaz, Mustansar and Debary, Hiyam and Fraccaro, Paolo and Paudel, Danda and Van Gool, Luc and Khan, Fahad and Khan, Salman},
  journal={arXiv preprint arXiv:2509.25026},
  year={2025}
}

@article{shabbir2025thinkgeo,
  title={Thinkgeo: Evaluating tool-augmented agents for remote sensing tasks},
  author={Shabbir, Akashah and Munir, Muhammad Akhtar and Dudhane, Akshay and Sheikh, Muhammad Umer and Khan, Muhammad Haris and Fraccaro, Paolo and Moreno, Juan Bernabe and Khan, Fahad Shahbaz and Khan, Salman},
  journal={arXiv preprint arXiv:2505.23752},
  year={2025}
}

@article{luo2025geoevolve,
  title={GeoEvolve: Automating Geospatial Model Discovery via Multi-Agent Large Language Models},
  author={Luo, Peng and Lou, Xiayin and Zheng, Yu and Zheng, Zhuo and Ermon, Stefano},
  journal={arXiv preprint arXiv:2509.21593},
  year={2025}
}

@article{wu2026vision,
  title={Vision-Language Reasoning for Geolocalization: A Reinforcement Learning Approach},
  author={Wu, Biao and Fang, Meng and Chen, Ling and Xu, Ke and Cheng, Tao and Wang, Jun},
  journal={arXiv preprint arXiv:2601.00388},
  year={2026}
}

@article{zhang2026instruction,
  title={Instruction tuning for large language models: A survey},
  author={Zhang, Shengyu and Dong, Linfeng and Li, Xiaoya and Zhang, Sen and Sun, Xiaofei and Wang, Shuhe and Li, Jiwei and Hu, Runyi and Zhang, Tianwei and Wang, Guoyin and others},
  journal={ACM Computing Surveys},
  volume={58},
  number={7},
  pages={1--36},
  year={2026},
  publisher={ACM New York, NY}
}

@inproceedings{chou2026test,
  title={Test-Time Consistency in Vision Language Models},
  author={Chou, Shih-Han and Chandhok, Shivam and Little, James J and Sigal, Leonid},
  booktitle={Proceedings of the IEEE/CVF Winter Conference on Applications of Computer Vision},
  pages={7789--7798},
  year={2026}
}

@article{yao2026remoteagent,
  title={RemoteAgent: Bridging Vague Human Intents and Earth Observation with RL-based Agentic MLLMs},
  author={Yao, Liang and Xu, Shengxiang and Liu, Fan and Zhang, Chuanyi and Yao, Bishun and Min, Rui and Li, Yongjun and Ouyang, Chaoqian and Di, Shimin and Zhang, Min-Ling},
  journal={arXiv preprint arXiv:2604.07765},
  year={2026}
}

@article{li2024language,
  title={Language-guided progressive attention for visual grounding in remote sensing images},
  author={Li, Ke and Wang, Di and Xu, Haojie and Zhong, Haodi and Wang, Cong},
  journal={IEEE Transactions on Geoscience and Remote Sensing},
  volume={62},
  pages={1--13},
  year={2024},
  publisher={IEEE}
}

@article{li2024show,
  title={Show me what and where has changed? question answering and grounding for remote sensing change detection},
  author={Li, Ke and Dong, Fuyu and Wang, Di and Li, Shaofeng and Wang, Quan and Gao, Xinbo and Chua, Tat-Seng},
  journal={arXiv preprint arXiv:2410.23828},
  year={2024}
}

@inproceedings{li2026rsvg,
  title={Rsvg-zeroov: Exploring a training-free framework for zero-shot open-vocabulary visual grounding in remote sensing images},
  author={Li, Ke and Wang, Di and Wang, Ting and Dong, Fuyu and Zhang, Yiming and Zhang, Luyao and Wang, Xiangyu and Li, Shaofeng and Wang, Quan},
  booktitle={Proceedings of the AAAI Conference on Artificial Intelligence},
  volume={40},
  number={8},
  pages={6288--6296},
  year={2026}
}

@article{li2026provg,
  title={ProVG: Progressive Visual Grounding via Language Decoupling for Remote Sensing Imagery},
  author={Li, Ke and Wang, Ting and Wang, Di and Zhu, Yongshan and Zhang, Yiming and Lei, Tao and Wang, Quan},
  journal={arXiv preprint arXiv:2604.01893},
  year={2026}
}
}


\end{document}